\documentclass[nohyperref]{article}

\usepackage{graphicx}
\usepackage{microtype}
\usepackage{subcaption}
\usepackage{booktabs} 

\usepackage{times}
\usepackage{latexsym}
\usepackage[utf8]{inputenc}
\usepackage{xcolor}
\usepackage{bm}
\usepackage{amsmath}
\usepackage{amssymb}
\usepackage{mathtools}
\usepackage{amsthm}

\usepackage{hyperref}

\usepackage[accepted]{icml2022}

\theoremstyle{plain}

\theoremstyle{definition}

\theoremstyle{remark}

\icmltitlerunning{Zero-Shot Transfer in Imitation Learning}
\begin{document}
    \twocolumn[
        \icmltitle{Zero-Shot Transfer in Imitation Learning}

        \icmlsetsymbol{equal}{*}

        \begin{icmlauthorlist}

            \icmlauthor{Alvaro Caudéran}{equal,eth-comp}
                        \icmlauthor{Gauthier Boeshertz}{equal,eth-comp}
            \icmlauthor{Florian Schwarb}{equal,eth-math}
            \icmlauthor{Calvin Zhang}{equal,eth-comp}
        \end{icmlauthorlist}
        \icmlaffiliation{eth-comp}{Department of Computer Science, ETH, Zürich, Switzerland}
        \icmlaffiliation{eth-math}{Department of Mathematics, ETH, Zürich, Switzerland}
        \icmlcorrespondingauthor{Alvaro Caudéran}{acauderan@ethz.ch}
    
        \icmlkeywords{Transfer Learning, Imitation Learning, Reinforcement Learning, Disentanglement Learning, Representation Learning}
        
        \vskip 0.3in
    ]
    
    \printAffiliationsAndNotice{\icmlEqualContribution}
    
    \begin{abstract}
            We present an algorithm that learns to imitate expert behavior and can transfer to previously unseen domains without retraining. Such an algorithm is extremely relevant in real-world applications such as robotic learning because 1) reward functions are difficult to design, 2) learned policies from one domain are difficult to deploy in another domain and 3) learning directly in the real world is either expensive or unfeasible due to security concerns. To overcome these constraints, we combine recent advances in Deep RL by using an AnnealedVAE to learn a disentangled state representation and imitate an expert by learning a single Q-function which avoids adversarial training. We demonstrate the effectiveness of our method in 3 environments ranging in difficulty and the type of transfer knowledge required. 
    \end{abstract}

    \section{Introduction} \label{sec:Introduction}
    Deep Learning has sparked success in a wide variety of previously challenging Reinforcement Learning (RL) domains \cite{DQN, alpha-fold, alpha-tensor}. Although its applications seem endless, some fundamental problems still remain unsolved. These include the availability of data, the lack of model interpretability \cite{towards-deep-symbolic-reinforcement-learning, elements-of-causal-inference}, the susceptibility of learned policies to changes in the input distribution, and the challenge to leverage expert data. In this paper, we combine recent advances in the latter two open problems. 
    
    \textbf{Domain adaptation}, a form of transfer learning, is the ability of RL agents to adapt to changes in the input distribution \cite{Bengio2012RepresentationLA}. In such a scenario, an RL agent is trained on a particular input distribution (\textit{source domain}) and is then placed in a setting where the input distribution is modified (\textit{target domain}). In many real-world applications, data from the target domain may be expensive, difficult to obtain, or not available at all \cite{generalizing-skills-with-semi-supervised-RL}. However, learning a policy by simply leveraging information from the source domain (called zero-shot learning) can lead to over-fitting to the input distribution, which results in poor adaptation performance \cite{building-machines-that-learn-and-think-like-people}. Therefore, it is crucial to learn a good low-dimensional state representation that is not task or domain-specific. There is a plethora of work that tries to learn a low-dimensional factorized state representation, which is called disentangled representation learning \cite{Schmidhuber, transformation-properties-of-learned-visual-representations, DC-IGN, BetaVAE_Kingma-Welling, CURL-2020, LUSR-2021}. A disentangled representation is defined as a factorized latent representation where either a single factor or a group of factors is responsible for the variation observed while it is invariant to changes in other factors \cite{Bengio2012RepresentationLA}. 
    
    \textbf{Imitation Learning (IL)} is the problem of learning to perform a task from expert trajectories. Approaches to IL can broadly be classified into two categories: 1) Behavioral Cloning (BC) \cite{behavioral-cloning-2010} or 2) Inverse Reinforcement Learning (IRL) \cite{ng2000algorithms}. BC is conceptually simple as it formalizes the IL problem as a supervised learning problem where the policy is a learned map between input states and output actions. This often requires a large number of trajectories \cite{ALVINN-1988} and small errors compound quickly. 
    A natural extension is to frame the IL problem as an inverse RL (IRL) problem: first, learn a reward function under which the expert's trajectories are optimal and from which a learned imitation policy can be trained \cite{gail-2016}. Much of the difficulty with this approach however relies on the min-max problem formulation over reward and policy. Instead, one can also learn a single model for the Q-value which implicitly defines both a reward and a policy function \cite{IQ-Learn-2021}. 
    
    A real-world application, where both the need for domain adaptation and imitation learning is evident, is the control of a robot arm trying to pick up an object. Firstly, training in the real world is slow and expensive as the robot might break and the repairs are costly. Second, it is hard to define a good reward function but generating a few expert trajectories manually can be trivial. Using imitation learning (IL) one can learn an optimal policy given the expert demonstrations in a simulation and then transfer it to the real world.
    
    There is some previous work that tries to combine IL and zero-shot transfer learning. However, most of these approaches rely either on the inferior behavioral cloning approach to IL \cite{Visual-Imitation-Made-Easy_2020, BC-Z-2022} or they use the complicated adversarial min-max approach to IRL \cite{ADAIL-2020}. Motivated by those deficiencies, we propose to use the approach of DARLA \cite{darla-2017} to learn a disentangled latent space representation of each state to adapt to changes in the input distribution. Using such latent representation, we aim to solve the problem of reward function definition by applying the IL approach IQ-Learn \cite{IQ-Learn-2021} to learn an optimal policy given expert demonstrations. 

    \section{Models and Methods}
    \subsection{Background}
    In this section, we briefly review the main concepts of the DARLA \cite{darla-2017} and the IQ-Learn \cite{IQ-Learn-2021} frameworks.
    
    \vspace{-0.1cm}
    \textbf{DARLA} (DisentAngled Representation Learning Agent) aims to perform zero-shot transfer learning to learn source policies $\pi_S$ that are robust to domain adaptation scenarios. The pipeline consists of the following three steps: 
    
    1) [Learn to see] The agent learns a mapping between the pixel observation state space and the latent state space. For this mapping, DARLA uses a $\beta$-VAE, which is a modification of the variational autoencoder \cite{BetaVAE_Kingma-Welling} that introduces an additional hyperparameter $\beta$ to balance reconstruction accuracy with latent channel capacity and independence constraints. For input pixels $x$ and latent representation $z$ in the source domain, it maximizes the objective function 
    \begin{equation}
        \begin{aligned}
            \mathcal{L}(\theta, \phi; x, z, \beta) = &\mathbb{E}_{z\sim q_{\phi}(\cdot \mid x)} \left[\log p_{\theta}(x \mid z)\right] \\
            & - \beta D_{KL}\left(q_{\phi}(z \mid x) \mid\mid p_{\theta} (z)\right). 
        \end{aligned}
        \label{eq:beta-VAE}
    \end{equation}
    The parameter $\phi$ parameterizes the decoder $q_{\phi}(z \mid x)$ and $\theta$ parametrizes the encoder $p_\theta(x \mid z)$ respectively. 
    
    2) [Learn to act] Using a standard RL algorithm, the agent is tasked to learn a source policy $\pi_S(a \mid z_S)$ based on the latent factors $z_S$ in the source domain $S$.
    
    3) [Transfer] The source policy $\pi_S$ is transferred to a target domain $T$ that the agent has previously not seen. There, the agent tries to collect rewards using as input state space the latent representations $z_T \sim p_{\phi}(\cdot \mid x_T)$ from the target domain $T$. 

    
    \vspace{-0.1cm}
    \textbf{IQ-Learn} (Inverse soft-Q Learning) proposes an elegant solution to the max-min problem of IRL by learning a single Q-value from which both the reward and policy function can be deduced. Given an expert policy $\pi^{e}$, the inverse reinforcement problem (IRL) tries to find a reward function $r$ that assigns a low cost to the expert policy and a high cost to any other policy $\pi$, i.e.,
    \begin{equation}
        \begin{aligned}
            \max_{r \in \mathcal{R}} \min_{\pi\in\Pi} L(\pi, r)
            =\max_{r\in \mathcal{R}} & \Big( \min_{\pi\in\Pi} \mathbb{E}_\pi\left[r(s, a)\right] - H(\pi) \Big) \\ 
            &- \mathbb{E}_{\pi^{e}} \left[r(s, a)\right] - \psi(r), 
        \end{aligned}
        \label{eq:IRL} 
    \end{equation}
    where $H(\pi) = \mathbb{E}_{\pi}[-\log \pi(a\mid s)]$ is the causal entropy, $\psi: \mathbb{R}^{\mathcal{R}} \to \mathbb{R} \cup \{\infty\}$ is a convex regularizer and $\mathcal{R}$ and $\Pi$ are the reward and policy space respectively. 
    
    It was then shown that \eqref{eq:IRL} can be equivalently written as
    \begin{equation}
        \max_{r \in \mathcal{R}} \min_{\pi\in\Pi} L(\pi, r) = \max_{Q \in \Omega} \mathcal{J^{*}}(Q), 
        \label{eq:transformed_objective}
    \end{equation}
    where 
    \begin{equation}
        \begin{aligned}
        \mathcal{J}^{*}(Q) &\triangleq  \mathbb{E}_{\rho_{e}}\left[\phi\big(Q(s, a) - \gamma \mathbb{E}_{s^{'} \sim \mathcal{P}(\cdot \mid s, a)} [V^{*}(s')]\big)\right] \\
        & -\mathbb{E}_{(s, a)\sim \mu} \left[V^{\pi}(s) - \gamma \mathbb{E}_{s'\sim\mathcal{P}(\cdot \mid s, a)}[V^{\pi}(s')]\right] 
        \end{aligned}
        \label{eq:Q_loss}
    \end{equation}
    for some concave function $\phi: \mathcal{R} \to \mathbb{R}$, occupancy measures $\rho_{e}$ and $\mu$ and $V^{*}(s) = \log \sum_{a'} \exp Q(s, a')$. An occupancy measure can be understood as the stationary distribution over $\mathcal{R}$ when running the policy $\pi$ in the environment. The objective function $\mathcal{J}^{*}$ is concave and can be easily optimized by gradient descent updates. Once an optimal $Q^{*}$ function was obtained from \eqref{eq:transformed_objective}, the corresponding reward $r^{*}$ and optimal policy $\pi^{*}$ satisfy
    \begin{equation}
        r^{*}(s, a) = Q^{*}(s, a) - \gamma \mathbb{E}_{s'\sim \mathcal{P}(\cdot |s, a)}\big[V^{*}(s')\big]
        \label{eq:optimal_reward}
    \end{equation}
    \begin{equation}
        \pi^{*}(a|s) = \frac{\exp Q^{*}(s, a)}{\sum_{a'} \exp Q^{*}(s, a')}.
        \label{eq:optimal_policy}
    \end{equation}

    \begin{figure*}[t!]
    \centering
        \includegraphics[width = 0.7\textwidth]{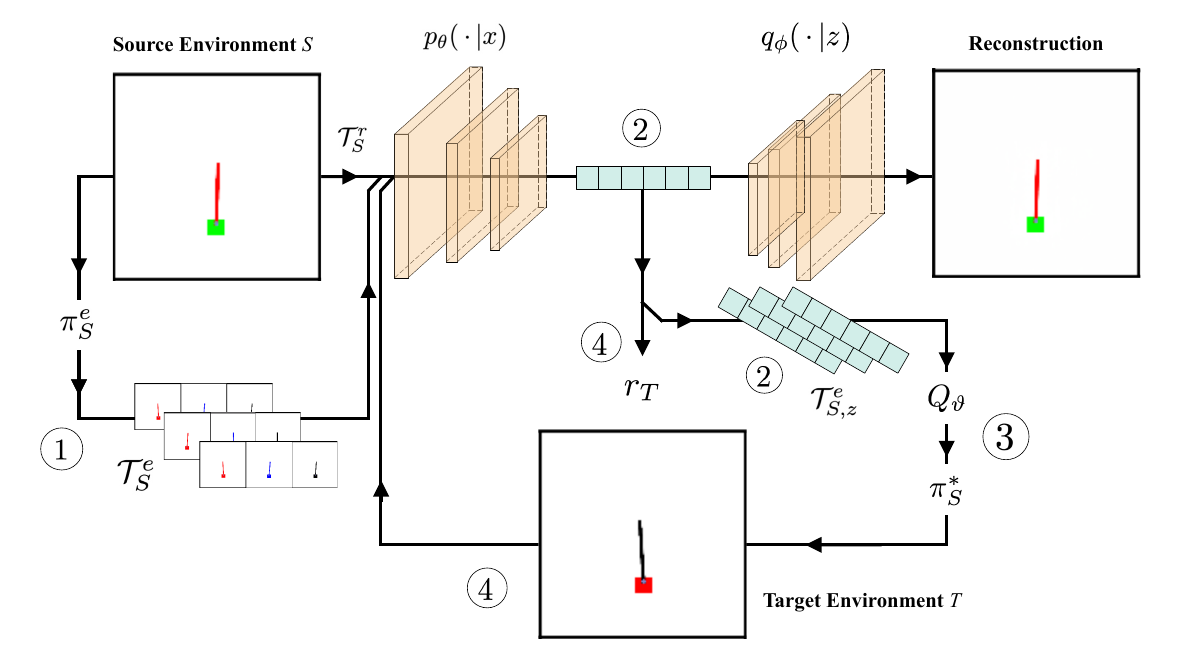}
        \caption{Visualization of our pipeline displaying the 4 intermediate steps}
        \label{fig:pipeline-structure}
    \end{figure*}
    
    \subsection{Algorithm}
    
    Our pipeline consists of the following 4 steps (implementation details can be found in Appendices \ref{appendix:vae}, \ref{appendix:ppo} and \ref{appendix:iq}): 
    
    1) Using a PPO-agent \cite{schulman2017proximal}, we learn an expert policy $\pi_{S}^{e}(a \mid x_S)$ in the source domain $S$. This is in contrast to DARLA, where the policy is being learned in the latent space directly. From this expert, we create trajectories $\mathcal{T}_S^{e}$, on which we base the imitation.
    
    2) We train an AnnealedVAE \cite{burgess2018understanding} based on some random  trajectories $\mathcal{T}_{S}^{r}$ in the source domain. In contrast to the $\beta$-VAE, the objective function of AnnealedVAE contains an additional constant $C\in \mathbb{R}$ and is given by
    \begin{equation}
        \begin{aligned}
            \mathcal{L}(\theta, \phi; x, z, \beta) & = \mathbb{E}_{z\sim q_{\phi}(z \mid x)} \left[\log p_{\theta}(x \mid z)\right] \\
            & - \beta \lvert D_{KL}\left(q_{\phi}(z \mid x) \, \mid\mid \, p_{\theta} (z)\right) - C\rvert. 
        \end{aligned}
        \label{eq:AnnealedVAE}
    \end{equation}
    By gradually adding more latent encoding capacity (i.e., increasing $C$) while training, we end up with more robust latent representations. We then encode the expert trajectory $\mathcal{T}_{S, z}^{e} \sim p_\theta(\cdot | \mathcal{T}_{S}^{e})$. 
    
    3) Motivated by the theoretical results in \eqref{eq:transformed_objective}, \eqref{eq:optimal_policy} and \eqref{eq:optimal_reward} we can imitate an expert $\pi^{e}$ using its encoded trajectories $\mathcal{T}_{S, z}^{e}$ in the latent space of the source domain $S$. 
    \begin{algorithm}
        \caption{Offline Inverse soft Q-Learning}
        \label{alg:offline-inverse-soft-q-learning}
        \begin{algorithmic}[1]
            \STATE {\bfseries Input:} $\mathcal{T}_{S, z}^{e}$, $N$
            \STATE Initialize Q-function $Q_{\vartheta}$
            \FOR{$t=1$ {\bfseries to} $N$}
                \STATE Train $Q$-function using objective from \eqref{eq:Q_loss}
                \STATE $\theta_{t+1} \gets \theta_t + \alpha \nabla_\vartheta \mathcal{J}(Q_{\theta})\vert_{\vartheta = \vartheta_t}$
            \ENDFOR
            \STATE Recover optimal policy: $\pi_S^{*} \gets \frac{1}{Z} \exp (Q_{\vartheta_N})$
        \end{algorithmic}
        \textbf{return} $\pi_S^{*}$ 
    \end{algorithm}
    
    Notice that $\mathcal{J}^{*}$ is defined \eqref{eq:Q_loss} for any occupancy measure $\mu$ and any concave function $\phi$. Thus, we can use the approximation
    \begin{align*}
        & \mathbb{E}_{(s, a)\sim \mu} \big[V^{\pi}(s) - \gamma \mathbb{E}_{s'\sim\mathcal{P}(\cdot | s, a)}[V^{\pi}(s')]\big] \\
        & \approx \mathbb{E}_{(s, a, s')\sim \text{expert}} \big[V^{\pi}(s) - \gamma V^{\pi}(s')\big] 
    \end{align*}
    and we keep $\phi(x)\triangleq x - \frac{1}{4\alpha} x^2$ as in the original paper. 

    \begin{table*}[t!]
        \caption{Zero-shot performance the \texttt{Cartpole Combi}, \texttt{Cartpole Background}, and \texttt{Super Mario} environments respectively. The mean and the standard deviation of the rewards collected for 20 different episodes are reported.}
        \label{tab:results}
        \centering
        \begin{tabular}{l | c c c}
             & & \textbf{Environments} & \\
             \textbf{Agents} & \texttt{Cartpole Combi} & \texttt{Cartpole Background} & \texttt{Super Mario} \\
             \hline
             \texttt{Random} & $21.42 \pm 11.80$ & $21.42 \pm 11.80$ & $353.46 \pm 137.56$ \\
             \texttt{PPO-Source} & $ 452.42 \pm 102.05 $ & $430.10 \pm 120.12$ & $2170.00 \pm 721.57$ \\
             \texttt{PPO-Target}& $449.77 \pm 110.05$ & $435.40 \pm 128.5$  & $1726.00 \pm 0.00$ \\  
             \texttt{PPO-Transfer} & $55.10 \pm 32.33 $ & $60.15 \pm 26.35$ & $556.00 \pm 0.00$ \\
             \texttt{PPO-Source-IQ} & $500.0 \pm 0.00 $& $467.90 \pm 87.06$  & $2342.33 \pm 798.70$ \\
             \texttt{PPO-Target-IQ} &$ 500.0 \pm 0.00$ & $458.30 \pm 75.1$ & $1726.00 \pm 0.00$ \\
             \hline 
             \textbf{Our Agent}  & $500.0 \pm 0.00$ & $325.50 \pm 119.00$ & $306.48 \pm 75.44$ \\
             \hline
        \end{tabular}
    \end{table*}
    
    4) We use the obtained policy $\pi_S^{*}$ and transfer it to the target domain $T$ without further retraining (zero-shot). 
    
    \subsection{Environments}
    We test our algorithm in 3 different environments that differ in the difficulty of the task, the type of transfer knowledge required, and/or the complexity of the latent space.  

    The Cartpole environment \cite{Gym-OpenAI} is a classic RL environment that we modified such that the colors of the background, the cart, and the pole can be changed. We exploit this in two ways: First, we consider combinations of different cart and pole colors. The target domain consists of a previously unseen combination of the cart and background color. Secondly, we change only the background color and transfer it to a previously unseen background color. We call the first modification \texttt{Cartpole Combi} and the latter \texttt{Cartpole Background}.
    
    Finally, we use the \texttt{Super Mario} environment \cite{gym-super-mario-bros}. This jump-and-run game comes with different levels, where the transfer consists of playing a previously unseen level. The choice of environments was guided by similarity in game objects, so the state distribution would not shift too much.
    
    The transitions used to train the encoder are gathered using a random agent for both environments. Appendices \ref{appendix:cartpole} and \ref{appendix:mario} present more details on the Cartpole and Super Mario environments.
    
    \section{Results}
    We compare our agent based on the rewards collected to the following baselines:
    \vspace{-0.1cm}
    \begin{enumerate}
        \vspace{-0.1cm}
        \item[1)] A random agent (\texttt{Random}),
        \vspace{-0.1cm}
        \item[2)] A PPO-agent trained on the source environment $S$ (\texttt{PPO-Source}),
        \vspace{-0.1cm}
        \item[3)] A PPO-agent trained on the target environment $T$ (\texttt{PPO-Target}), 
        \vspace{-0.1cm}
        \item[4)] An agent trained on the source environment $S$ and directly transferred to the target environment without using the AnnealedVAE. (\texttt{PPO-Transfer}),
        \vspace{-0.1cm}
        \item[5)] An PPO-agent trained on the source environment $S$ and imitated by Algorithm \ref{alg:offline-inverse-soft-q-learning} (\texttt{PP0-Source-IQ}),
        \vspace{-0.1cm}
        \item[6)] An PPO-agent trained on the target environment $T$ and imitated by Algorithm \ref{alg:offline-inverse-soft-q-learning} (\texttt{PPO-Target-IQ}).
    \end{enumerate}
    \vspace{-0.1cm}
    Each of these baseline agents as well as our own agent are then allowed to collect rewards in the environment for 500 steps in the Cartpole environments and 4500 steps in the Super Mario environment respectively. The mean reward and the standard deviation over 20 different episodes with different starting states are reported in Table \ref{tab:results}. We report the best evaluation scores over the training run and the transfer score with the model that achieved the best evaluation.

    
    \section{Discussion} \label{sec:discussion}
    Let us first look at the transfer alone. By comparing the \texttt{Random} agent with the \texttt{PPO-Transfer} agent, we see that without using the AnnealedVAE, the transfer of a PPO policy to a previously unobserved target environment performs barely better than a random policy. This holds true for all 3 environments and shows that transferring policies without further retraining (zero-shot) is nearly impossible.

    Now let us compare our agent to the baselines. For \texttt{Cartpole Combi} we see that for any of the 20 episodes we achieved maximum reward. This means that the combination of transfer and imitation worked really well. For the second environment, \texttt{Cartpole Background}, we perform slightly worse than a PPO agent trained on the target environment (\texttt{PPO-Target}) and its imitation (\texttt{PPO-Target-IQ}). Nonetheless, we perform significantly better than both a \texttt{Random} policy or a PPO-agent that was transferred without using the AnnealedVAE. Moreover, the imitated expert for \texttt{Cartpole Background} did in both domains not reach optimal rewards. This is another proof that the AnnealedVAE gives too much weight on the background color and loses focus on the cart and the pole. If we now move to our reach task, the \texttt{Super Mario} environment, we see that our agent does not outperform a random agent. Indeed, our agent does not reach reward levels from either plain or imitated PPO agents in any environment. As the imitated agents (\texttt{PPO-Source-IQ} and \texttt{PPO-Target-IQ}) reach comparable rewards to their expert counterparts, we conclude that difficulty comes from the transfer. This is underlined by the \texttt{PPO-Transfer} agent which already achieves only a slightly better reward than a random agent. The transfer here is harder than in the Cartpole environments because in the latter the state space is similar up to colors, whereas the transfer in Super Mario also changes the state of the environment.

    This shows that learning a disentangled representation of the feature space is extremely challenging and not possible for arbitrary shifts in the input distribution as the \texttt{Super Mario} environment showed. We found that training an AnnealedVAE on multiple versions of the environment like multiple background colors or multiple levels as in Super Mario helps to detect relevant latent factors. 
    
    \textbf{Challenges and Further Work} \\
    The challenges of learning a disentangled state representation are well-known \cite{challenges-in-disentanglement-learning}. For example, for Gaussian latent factors, one cannot distinguish between any rotations applied to the latent factors. Moreover, there seem to be no clear choices of latent dimensions, and pruning latent factors did not help disentangle the features. As an outlook, it would be interesting to see how one(or even multi)-shot transfer performs. 
        
    \section{Summary} \label{sec:summary}
    We conclude that we were able to perfectly zero-shot transfer an imitated expert policy for \texttt{Cartpole Combi} environment. Transferring to environments with previously unobserved distribution shifts such as \texttt{Cartpole Background} or even different states as in \texttt{Super Mario} is highly non-trivial. Nevertheless, expert imitation works still very well in these challenging environments. 


\newpage
\bibliography{bibliography}
\bibliographystyle{icml2022}

\newpage
\appendix
\onecolumn

\section{AnnealedVAE} \label{appendix:vae}
    For all our experiments, we used a $128 \times 128 \times 3$ dimensional state space, where we stacked 4 observations into one image. 
    
    The encoder and decoder of the AnnealedVAE consisted of 4 convolutional layers of kernel sizes $\{$32, 64, 128, 256$\}$ as well as 2 fully connected layers of size $1028$ for $\mu$ and $\log(\sigma^2)$. Moreover, we used $\beta = 4$ and a latent state space size of $10$ for the Cartpole environments and of size $32$ for the Super Mario environment. We use the Adam optimizer \cite{kingma2014adam} with a learning rate of $1 \times 10^{-4}$. The output of each convolutional layer goes through a group norm layer with group size 32, it then goes through LeakyReLU non-linearities. The difference between the AnnealedVAE and the standard $\beta$-VAE is that the KL divergence term in the loss function is constrained by subtracting a value C from it. This value C starts low and finishes high such that the KL loss will not impact the loss at the end of training to provide good reconstruction thanks to the reconstruction loss being larger than the KL loss. The maximum value for this C is 25 for all environments.

\begin{figure}[h]
    \caption{Traversal of latent space. Each row is one of the 10 latent factors.}
    \centering
    \includegraphics[width=0.5\textwidth]{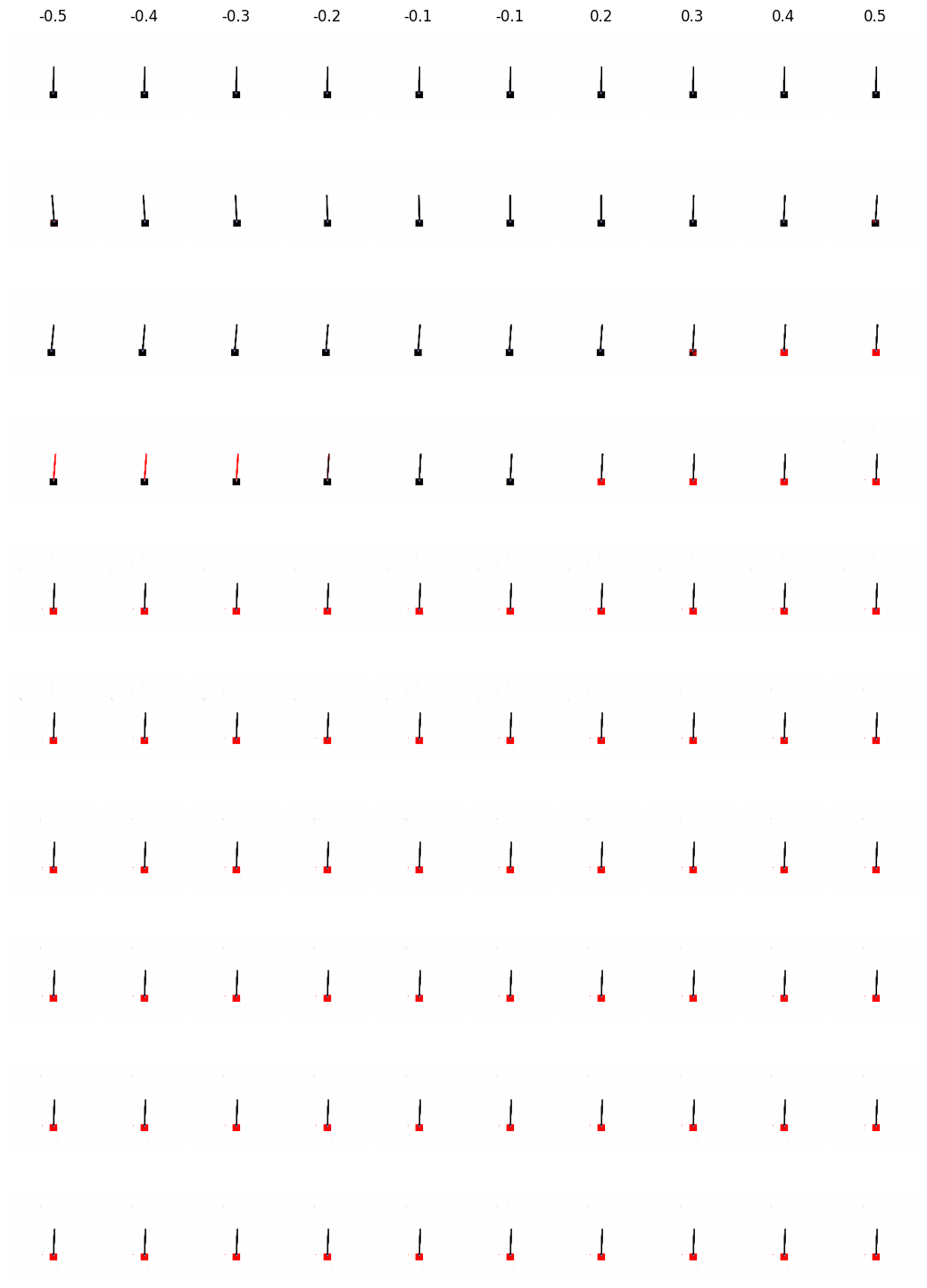}
    \label{fig:bvae_disent}
    \end{figure}

    As discussed in \cite{challenges-in-disentanglement-learning}, there are no guarantees that a $\beta$-VAE will properly disentangle the factors in an image. This is also the case for the AnnealedVAE we use for \texttt{Cartpole Combi}. We show a latent traversal in figure \ref{fig:bvae_disent}. We can see that the 3rd and 4th rows both change the colors of the cart and pole, so we can say that the factor representing the colors is not properly disentangled. However, we think that perfect disentanglement is not required as the transfer still works, that is, the Soft-Q network can still get enough information from the latent encodings even though they are entangled.

\section{Expert Agent Training with PPO} \label{appendix:ppo}

    In order to train the experts for generating trajectories, we make use of the Stable Baselines 3 package \cite{stable-baselines3}. In particular, we use PPO \cite{schulman2017proximal} as a learning algorithm parameterized by a neural network with the same architecture as in \cite{mnih2015human}.

    For the Cartpole environments, we train for $10^6$ time steps with a batch size of $128$ and a learning rate of $1 \times 10^{-4}$ with a linear decay. The rest of the parameters for PPO are the default ones. We also use an upper bound of 500 to stop training as the agent would otherwise be allowed to balance the pole indefinitely.

    For the Super Mario environment, the training is performed on a vectorized environment containing observations from all training levels (more details on the environment in Appendix \ref{appendix:mario}). In this setting, we train for $5 \times 10^6$ time steps with a batch size of 32 and an exponentially decaying learning rate starting from $2.5 \times 10^{-4}$. We also change the coefficient of the value function to $0.5$ and the one for entropy to $0.01$ for PPO.

\section{IQ Learn} \label{appendix:iq}
    \begin{figure}[h]
    \caption{Average rewards of the evaluation on Cartpole against the number of expert trajectories, the shaded area is the standard deviation of the trials}
    \centering
    \includegraphics[width=0.5\textwidth]{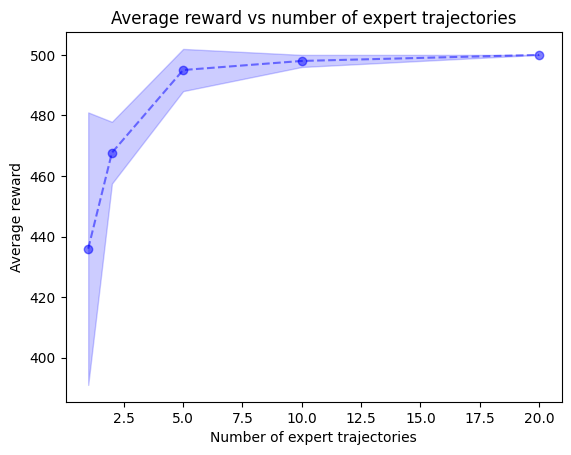}
    \label{fig:rew_vs_num_traj}
    \end{figure}

    To train the imitation learning algorithm, we used a network of two layers of size 64 for every environment, trained with the Adam optimizer \cite{kingma2014adam} and a learning rate of $1 \times 10^{-4}$ for $10^5$ steps with the $\chi$-square added to the loss The other parameters include $\gamma = 0.99$ and $\alpha = 0.5$.
    
    We used IQ Learn due to its efficiency with respect to expert trajectories. To demonstrate this we plotted the average rewards of evaluation trials of the recovered agent on Cartpole against the number of expert trajectories in figure \ref{fig:rew_vs_num_traj}. As we can see, the average reward is close to perfect with one trajectory and quickly plateaus to 500.

\section{Cartpole} \label{appendix:cartpole}
    Cartpole is a standard state-based RL environment. We modified it to output images such that we can easily modify the background, cart, and pole colors easily. We use two different configurations of Cartpole as explained above. The first is the \texttt{Cartpole Combi} and \texttt{Cartpole Background}.
    
    \subsection{Cartpole Combi}
    This configuration is similar to what is done in \cite{darla-2017} as we train on different combinations of colors and transfer to an unseen combination of colors. Specifically, we train on three different environments, the first is with a black cart and pole, the second is with a black pole and red cart, and finally with a red pole and black cart. We then transfer to a red cart and pole. This is intended to be an easier environment. 
    
    \subsection{Cartpole Background}
    This is a harder configuration where we train the AnnealedVAE on two colors only (i.e. (255, 255, 255) and (150, 202, 124)). We then recover the policy on an environment with one of these two colors (i.e. (255, 255, 255)). Finally, we transfer the recovered policy to an environment whose background color has never been seen (i.e. (215,148,187)), neither by the AnnealedVAE nor the agent. Interestingly, we observe that as the target color becomes more distanced from the trained colors, the transfer becomes worse. This makes sense, considering the latent space is not properly disentangled, compounding errors are to be expected.

\section{Super Mario Environment} \label{appendix:mario}

\begin{figure}[h]
    \begin{subfigure}{.25\textwidth}
      \centering
      \includegraphics[width=0.9\linewidth]{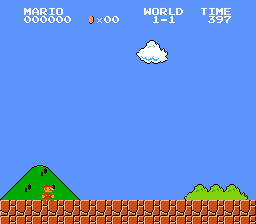}
      \caption{Standard}
    \end{subfigure}%
    \begin{subfigure}{.25\textwidth}
      \centering
      \includegraphics[width=0.9\linewidth]{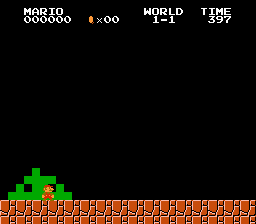}
      \caption{Downsampled}
    \end{subfigure}%
    \begin{subfigure}{.25\textwidth}
      \centering
      \includegraphics[width=0.9\linewidth]{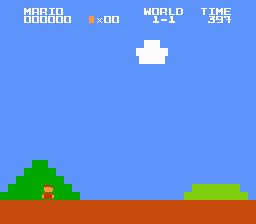}
      \caption{Pixel}
    \end{subfigure}%
    \begin{subfigure}{.25\textwidth}
      \centering
      \includegraphics[width=0.9\linewidth]{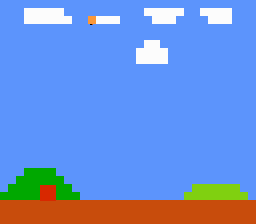}
      \caption{Rectangle}
    \end{subfigure}
    \caption{Different versions provided by the Super Mario environment.}
    \label{fig:mario-versions}
    \end{figure}

    The Super Mario environment \cite{gym-super-mario-bros} is an extension to the classical OpenAI Gym environments. In particular, it provides 8 worlds $\times$ 4 stages $=$ 32 levels in total. In addition, it provides 4 versions for each level: \texttt{standard}, \texttt{downsampled}, \texttt{pixel}, and \texttt{rectangle}. For the purpose of this paper, we chose to use the \texttt{rectangle} version of the levels, as it provides simpler frames to learn for the autoencoder. The difference between the four versions can be observed in Figure \ref{fig:mario-versions}.

\begin{figure}[h]
    \begin{subfigure}{.25\textwidth}
      \centering
      \includegraphics[width=.9\linewidth]{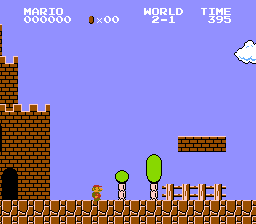}
      \caption{Level 2-1}
    \end{subfigure}%
    \begin{subfigure}{.25\textwidth}
      \centering
      \includegraphics[width=.9\linewidth]{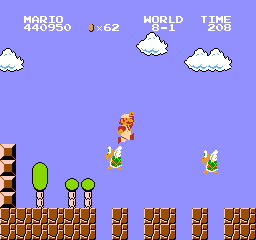}
      \caption{Level 8-1}
    \end{subfigure}%
    \begin{subfigure}{.25\textwidth}
      \centering
      \includegraphics[width=.9\linewidth]{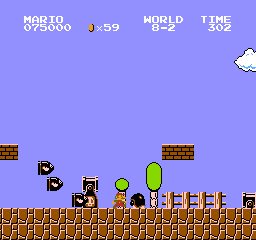}
      \caption{Level 8-2}
    \end{subfigure}%
    \begin{subfigure}{.25\textwidth}
      \centering
      \includegraphics[width=.9\linewidth]{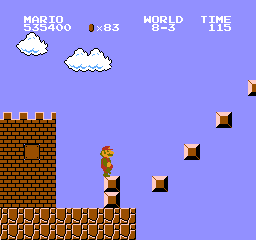}
      \caption{Level 8-3}
    \end{subfigure}
    \caption{Different levels of Super Mario used in this paper.}
    \label{fig:mario-levels}
    \end{figure}

    Let us denote with \texttt{(world number)}-\texttt{(stage number)} the level, where the first number is the world and the second is the stage. Then the training of the expert agent is done on levels 8-1, 8-2, and 8-3. We then perform the transfer to level 2-1. The levels were chosen so that the elements in the target level are a subset of the ones in the training world. A graphical representation of the similarity between the levels can be observed in Figure \ref{fig:mario-levels}.
    
\end{document}